\documentclass[conference]{IEEEtran}
\usepackage[utf8]{inputenc}
\usepackage[numbers]{natbib}
\usepackage{amsmath}
\usepackage{amsfonts}
\usepackage{graphicx}
\usepackage{booktabs}
\usepackage{tabularx}

\newcommand{\given}{\lvert}
\DeclareMathOperator{\Gauss}{\mathcal N}
\newcommand{\KL}[2]{D_{\text{KL}}\left(#1 \mid\mid #2\right)}
\DeclareMathOperator{\E}{\mathbb{E}}

\ifCLASSINFOpdf
\else
\fi
\makeatletter
\g@addto@macro\normalsize{%
  \setlength\abovedisplayskip{11pt}
  \setlength\belowdisplayskip{11pt}
  \setlength\abovedisplayshortskip{11pt}
  \setlength\belowdisplayshortskip{11pt}
}
\makeatother
\begin{document}
%\title{Using Synthetic Data to Train Neural Networks is Model-Based Reasoning: \\ Breaking Captcha without Labeled Data}
% \title{Synthetic Data and Model-Based Reasoning:\\ Breaking Captcha without Labeled Data}
\title{Using Synthetic Data to Train Neural Networks is Model-Based Reasoning}

\author{\IEEEauthorblockN{Tuan~Anh Le\IEEEauthorrefmark{1},
Atılım Güneş Baydin\IEEEauthorrefmark{1},
Robert Zinkov\IEEEauthorrefmark{2} and
Frank Wood\IEEEauthorrefmark{1}}
\IEEEauthorblockA{\IEEEauthorrefmark{1}Department of Engineering Science\\University of Oxford, Parks Road, OX1 3PJ Oxford, UK\\ Email: \{tuananh, gunes, fwood\}@robots.ox.ac.uk}
\IEEEauthorblockA{\IEEEauthorrefmark{2}School of Informatics and Computing\\Indiana University, 919 E 10th Street,
Bloomington, IN 47408, USA\\
Email: zinkov@iu.edu}}


\maketitle

\begin{abstract}
We draw a formal connection between using synthetic training data to optimize neural network parameters and approximate, Bayesian, model-based reasoning. In particular, training a neural network using synthetic data can be viewed as learning a proposal distribution generator for approximate inference in the synthetic-data generative model. We demonstrate this connection in a recognition task where we develop a novel Captcha-breaking architecture and train it using synthetic data, demonstrating both state-of-the-art performance and a way of computing task-specific posterior uncertainty. Using a neural network trained this way, we also demonstrate successful breaking of real-world Captchas currently used by Facebook and Wikipedia. Reasoning from these empirical results and drawing connections with Bayesian modeling, we discuss the robustness of synthetic data results and suggest important considerations for ensuring good neural network generalization when training with synthetic data.
\end{abstract}


% \fi
\IEEEpeerreviewmaketitle

\section{Introduction}
\label{sec:introduction}

Neural networks are powerful regressors \cite{lecun2015deep}. Training a neural network for regression means finding values for its free parameters using supervised learning techniques. This generally requires a large amount of labeled training data.  Generally the harder the task, the larger the neural network, and the more training data required.

When labeled training data are scarce, one must either generate and use synthetic data to train, or resort to unsupervised generative modeling and generally slow test-time inference since it must be run afresh for new data. The deep learning community has reported remarkable results taking the former approach, either in the limited form of data augmentation \cite{simard2003,krizhevsky2012imagenet}, where a dataset is artificially enlarged using label-preserving transformations, or training models solely on synthetic data, such as the groundbreaking work on text recognition in the wild \citep{jaderberg2014synthetic,jaderberg2016reading,gupta2016synthetic}, which was achieved by training a neural network to recognize text using synthetically generated realistic renders. \citet{GoodfellowBIAS13} addressed recognition of house numbers in Google Street View images in a supervised fashion, also solving reCaptcha \cite{vonahn2008recaptcha} images using synthetic data to train a recognition network from image to latent text. That the authors were Google employees meant that they had access to the true reCaptcha generative model and thus could generate millions of labeled instances for use in a standard supervised-learning pipeline. More recently, \citet{stark-gcpr15} also used synthetic data for Captcha-solving and \citet{wang2015deepfont} for font identification. 
 
A contribution of this paper is to point out that this kind of use of synthetic data to train a neural network under a standard loss is, in fact, equivalent to training an artifact to do amortized approximate inference, in the sense of \citet{gershman2014amortized}, for the generative model corresponding to the synthetic data generator. This relationship forms the basis of our recent work on inference compilation for probabilistic programming \cite{le2016inference} and is also noted by both \citet{paige2016inference} and \citet{papamakarios2016fast}, where approximate inference guided by neural proposals is the goal rather than training neural networks using synthetic data. A consequence of this is that there is no need to ever reuse training data, as ``infinite'' labeled training data can be generated at training time from the generative model. Another contribution we make is a suggestion for how to take advantage of this framework by running a neural network more than once at test time to compute task-specific uncertainties of interest.   

These contributions can also be seen as a reminder and guidance to the neural network community as it continues to move towards tackling unsupervised inference and problems in which labeled training data are difficult or impossible to obtain.  Towards this end, we examine experimental findings that highlight problems that are likely to arise when using synthetic data to train neural networks.  We discuss these problems in terms of the brittleness demonstrated to exist for deep neural networks, for example by \citet{szegedy2013intriguing}, who showed that perceptually indistinguishable variations in neural network input can lead to profound changes in output. We also discuss model misspecification in the Bayesian sense \cite{gelman2013philosophy}.

The paper structure is as follows. In Section~\ref{sec:captcha-breaking}, we develop a probabilistic synthetic data generative model and suggest a single, flexible neural network architecture for Captcha-breaking.  In Section~\ref{sec:experiments}, we train each such model independently using training data derived from running the synthetic data generator with parameters set to produce the corresponding style.  These neural networks are shown to produce extremely good breaking performance, both in terms of accuracy and speed, well beyond standard computer vision pipeline results and comparable to recent deep learning results. We then discuss and demonstrate the brittleness of these regressors.  We demonstrate improved robustness by focusing on and improving the generative model. In Section~\ref{sec:connections}, we illustrate the connection of the demonstrated brittleness with Bayesian model mismatch. We end by explaining how the learned neural network can be used to perform sample-based approximate inference.

\begin{table*}[t]
  \footnotesize
  \setlength{\tabcolsep}{1mm}
  \caption{Synthetic Captcha breaking results. RR: recognition rate, BT: breaking time.}
  \label{table:results/captcha-broken}
  \def\arraystretch{1.25}
  \begin{tabularx}{\textwidth}{@{}p{34mm}lllXXXXX@{}}
    \toprule
    Style & & Baidu (2011) & Baidu (2013) & eBay & Yahoo & reCaptcha & Wikipedia & Facebook \\
    & &  \includegraphics[width=18mm]{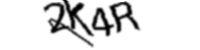} & \includegraphics[width=18mm]{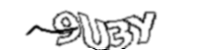} & \includegraphics[width=18mm]{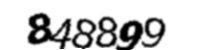} & \includegraphics[width=18mm]{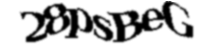} & \includegraphics[width=18mm]{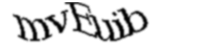} & \includegraphics[width=18mm]{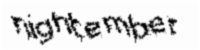} & \includegraphics[width=18mm]{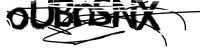} \\
    \midrule
    % Compiled inference & Training traces & 15.11M & 19.04M & 11.15M & 16.22M & 10.36M & 19.01M\\
    % & Initial loss & 50.11 & 49.58 & 147.35 & 70.57 & 71.59 & 76.52\\
    % & Final loss & 0.37 & 0.36 & 5.20 & 0.08 & 0.27 & 0.70\\
    Our method & RR & 99.8\% & 99.9\% & 99.2\% & 98.4\% & 96.4\% & 93.6\% & 91.0\%\\
    & BT & 72 ms & 67 ms & 122 ms & 106 ms & 78 ms & 90 ms & 90 ms\\
    \midrule
    \citet{bursztein2014end}& RR & 38.68\% & 55.22\% & 51.39\% & 5.33\% & 22.67\% & 28.29\% \\
    & BT & 3.94 s & 1.9 s & 2.31 s & 7.95 s & 4.59 s \\
    \midrule
    \citet{starostenko2015breaking}& RR & & & & 91.5\% & 54.6\% & \\
    & BT & & & & & $<$ 0.5 s\\
    \midrule
    \citet{gao2014robustness} & RR & 34\% & & & 55\% & 34\% & & \\
    \midrule
    \citet{gao2013robustness} & RR & & 51\% & & 36\% \\
    & BT & & 7.58 s & & 14.72 s \\
    \midrule
    \citet{GoodfellowBIAS13} & RR & & & & & 99.8\%\\
    \midrule
    \citet{stark-gcpr15} & RR & & & & & 90\% \\
    \bottomrule
\end{tabularx}
\end{table*}

\section{Captcha-breaking}
\label{sec:captcha-breaking}

Assuming no access to the true Captcha \cite{vonahn2003captcha} generating system and a paucity of labeled training data, how does one go about breaking Captchas?  A hint appears in the probabilistic programming community's approach to procedural graphics \cite{mansinghka2013approximate} where a generative model for Captchas is proposed and then general purpose Markov chain Monte Carlo (MCMC) Bayesian inference is used to computationally inefficiently invert the said model.  We will make the argument that this is, effectively, the same as generating synthetic training data in the manner of \citet{jaderberg2014synthetic,jaderberg2016reading} to train a neural network that regresses to the latent Captcha variables.  In either case, developing a flexible, well-calibrated synthetic training data generator is our first concern.

\subsection{Generating synthetic training data}

Our synthetic data generative model for Captcha specifies joint densities $p_s(x, y)$, parameterized by style $s$, that describe how to generate both the latent random variable $x$ and the corresponding Captcha image $y$.  Referring to the first row of Table~\ref{table:results/captcha-broken}, style $s$ pertains to different schemes (e.g., Baidu, eBay, Wikipedia, Facebook) involving distinct character ranges, fonts, kerning, deformations, and noise.  Note that in the following equations we omit the style subscript while keeping in mind that there is a separate unique model for each style.
The latent structured random variable $x = \{L,\epsilon_{1:K}, i_{1:L}\}$ includes $L$, the number of letters, $\epsilon_{1:K}$, a multidimensional structured parameter set controlling Captcha-rendering parameters such as kerning and various style-specific deformations, and $i_{i:L}$, letter identities.
Given these, we use a custom stochastic Captcha renderer $\mathcal{R}$ to generate each Captcha image $y$, this renderer and its fidelity being the primary component of the synthetic data generation effort.  The corresponding per-style synthetic data generator corresponds to the model
\begin{align}
    x &\sim p(x) \label{eq:algorithm/nn/prior}\\
    y|x &\sim \mathcal{R}(x)\;, \label{eq:algorithm/nn/render}
\end{align}
where $p(x)$ is a style-specific prior distribution over the latent variables including the character identities. For each different style shown in Table~\ref{table:results/captcha-broken}, we use different settings of the prior parameters to drive the Captcha renderer.  In particular, the model places style-specific uniform distributions over different intervals for $L$, $\epsilon_{1:K}$, and $i_{1:L}$.
This is the mechanism for generating synthetic training data $\{(x^{(n)}, y^{(n)})\}$.
Note that $p(y|x)$ cannot be evaluated for a given $y$, rather only sampled.

\subsection{Neural network architecture}

Our Captcha-breaking neural network is designed taking into account architectures that have been shown to perform well on image inputs and variable-length output sequences \cite{vinyals2015show,karpathy2015deep}. Specifically, we choose a combination of convolutional neural networks (CNNs) and recurrent neural networks.

\begin{figure}
    \centering
    \includegraphics[width=\linewidth]{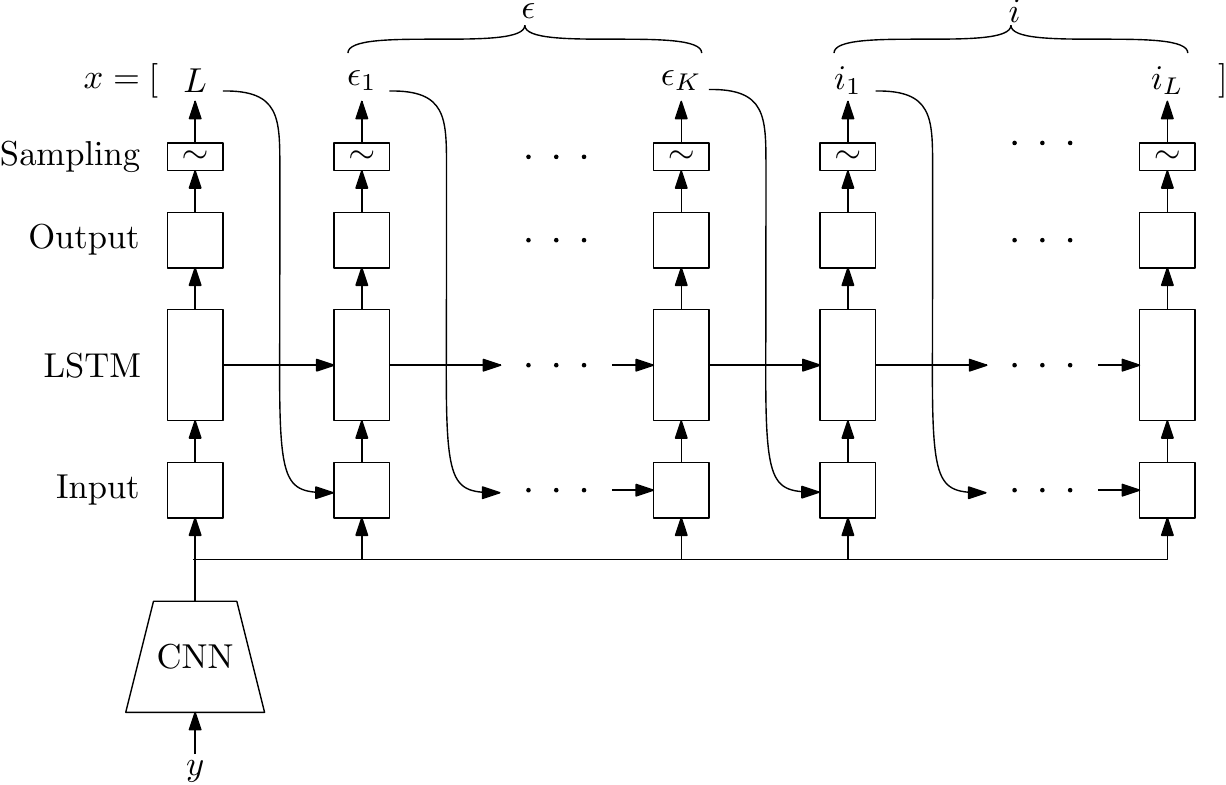}
    \caption{Neural network architecture mapping the Captcha image $y$ to the latent variables $x$ of interest.}
    \label{fig:algorithm/nn_arch}
\end{figure}

The core of our neural architecture (Figure~\ref{fig:algorithm/nn_arch}) is a long short-term memory (LSTM) network \cite{hochreiter1997long}, the output of which at each time step is passed through output layers corresponding one-to-one to the components of the latent variable $x$ in the generative model (i.e., number of letters $L$, rendering parameters $\epsilon_{1:K}$, and letter identities $i_{1:L}$) that constitute the inputs to the Captcha renderer. Since the latent variable $x$ has $T = 1 + K + L$ components, where $K$ is style-specific and $L$ is instance-specific, the LSTM is run for $T$ time steps, and we represent by $x_{1:T}$ the components of the latent $x$ at each time step. The output layers are fully-connected layers followed by a softmax function, distinct for each latent variable, that parameterize a discrete probability distribution. Since the LSTM has a fixed-dimensional output, these output layers allow us to match the dimensions of the discrete distributions for the corresponding latent variables.

A CNN is used to embed the Captcha image $y$ into a fixed-dimensional embedding vector CNN($y$). At each time step, the LSTM input is constructed as the concatenation of the image embedding CNN($y$), the value of the latent variable $x_{t-1}$ of the previous time step, and a label vector $\{0, 1\}^D$ corresponding to each $x_t$. During training, all $x_{1:T}$ are provided to the network in a way similar to that used by \citet{reed2015neural}, using the actual values that generated the synthetic image $y$. At test time, the values of $x_t$ are sampled from the corresponding discrete probability distribution.

We denote the combined set of parameters of the overall architecture $\theta$ and its forward propagation function $\eta$, so given an input $y$, the output of the softmax layer at time step $t$ corresponding to $x_t$ is $\eta_{\theta, t}(y)$.
In the running example of Figure~\ref{fig:algorithm/nn_arch}, $x_1 = L$, $x_{2:(2 + K - 1)} = \epsilon_{1:K}$, and $x_{(2 + K):(2 + K + L - 1)} = i_{1:L}$.

\subsection{Loss}
By design, the softmax outputs determine the parameters for the discrete probability distributions of the Captcha generator parameters.  The loss we minimize during training is the negative sum of the log of the softmax outputs
\begin{align}
    \mathcal L(\theta) &= \frac{1}{N}\sum_{n = 1}^N \left[-\sum_{t = 1}^T \log\left( [\eta_{\theta, t}(y^{(n)})]_{x_t^{(n)}} \right)\right]\;, \label{eq:nn/loss}
\end{align}
where we use the notation $[z]_i$ to denote the $i$th element of $z$.
This is a standard loss used in training neural networks for classification.
The connection with Bayesian modeling in which we interpret softmax outputs as probabilities of discrete random variables in a joint importance sampling proposal distribution is explored in more detail in Section~\ref{sec:connections/inference}.
% Since the softmax outputs can be interpreted as probabilities of discrete random variables, this minimization can be interpreted as being over the expectation of log joint probabilities. This connection is explored in more detail in Section~\ref{sec:connections/inference}.
% Note that this is equivalent to maximizing a product of probabilities.
%where $[z]_i$ is the $i$th element of $z$.

\section{Experiments}
\label{sec:experiments}

We wrote synthetic data generative models for seven different Captcha styles, covering the types frequently found in the Captcha breaking literature \cite{starostenko2015breaking,bursztein2014end,gao2013robustness,gao2014robustness}.  For each of these, we trained a neural architecture consisting of (1) a CNN with six convolutions (3$\times$3, with successively 64, 64, 64, 128, 128, 128 filters), max-pooling (2$\times$2, step size 2) after the second, fifth, and sixth convolutions, and two final fully-connected layers of 1024 units; (2) a stack of two LSTMs of 512 hidden units each; and (3) fully-connected layers of appropriate dimension mapping the LSTM output to the corresponding softmax dimension of each latent variable. ReLU activations were used after the convolutions and the fully-connected layers overall.

We empirically verified that supplying the image embedding CNN($y$) to the LSTM at every time step makes the training progress faster in our setup where we train the CNN from scratch together with the rest of the components, compared with the alternative of using CNN($y$) only once and pretraining CNN weights on an image recognition database as in \citet{vinyals2015show} and \citet{karpathy2015deep}.

% Note that at training time we have access to all elements of $x$; however, at test time we do not.
% Classification in our architecture is stochastic with a guided but stochastic choice made by sampling from the sequence of LSTM outputs in a way much
% closer to the MADE \cite{germain2015made}.

The networks were implemented in Torch \cite{collobert2011torch7} and trained with Adam \cite{kingma2015adam} optimization, with initial learning rate $\alpha = 0.0001$, hyperparameters $\beta_1 = 0.9$, $\beta_2 = 0.999$, using minibatches of size 128. The generative models were implemented in the Anglican probabilistic programming language \cite{wood2014new}. The two are coupled in our inference compilation \cite{le2016inference} framework.\footnote{https://probprog.github.io/inference-compilation/}

% Maybe (space permitting) add the ``infinite data'' figure and discuss.

\subsection{Initial results}

As can be seen in Table~\ref{table:results/captcha-broken}, this architecture, and our method for training it using synthetic data, outperforms nearly all state-of-the-art Captcha breakers in terms of both accuracy and recognition times with the exception of \citet{GoodfellowBIAS13}, which used data drawn from the true reCaptcha generator.
The row labeled ``our method'' shows breaking results and speeds for our neural network trained using synthetic data to decode unlabeled Captchas from the same Captcha generator.
The \citet{GoodfellowBIAS13} and \citet{stark-gcpr15} rows show the most directly comparable results, namely, using deep neural networks to break unlabeled Captchas training on synthetic data.
The additional rows show breaking results for more traditional segment-and-classify computer vision image processing pipelines.
These, in contrast to the others, do not have access to the true Captcha generator but instead report test results on real-world Captchas gathered in the wild.
If robust, $>90\%$ accuracies would seem to confirm that Captcha, from a computer security perspective \cite{bursztein2011text,sivakorn2016robot}, is indeed broken.

While the capabilities of deep neural networks are impressive, it should be noted that these kinds of results, on occasion, can be somewhat misleading \cite{szegedy2013intriguing}.  In particular, one should note the assumption that, up to this point in this paper and in the referenced results from the deep learning literature, the training procedure of the Captcha-breaking network has access to data from the true generative process.  Indeed, samples from the true generative process are superior even to hand-labeled training instances gathered in the wild.  Any simulated data, required when we do not have access to the true generative model, must come from an approximation to the true generative process, a model per se.  Whether or not networks trained using such approximate data are robust in the sense of working well on real data in the wild becomes the real question.  To put it another way, is Captcha really broken if we do not have access to the true generative model---or a legion of human labelers and a pile of cash?

\subsection{Robustness of results}
\label{sec:experiments/robustness}
So, what happens to these state-of-the-art models if the test data is subtly different to the generated synthetic data?  Or, what happens if you attempt to transfer learning from one Captcha style to another?  Our exploration of these questions forms the inspiration and basis for the rest of the paper.

To start, we tried to use our trained models on real Captchas from Wikipedia and Facebook, which we identified as two major web services that still make use of textual Captchas,\footnote{Facebook Captchas appear as a measure for preventing flood-posting and when links to particular Facebook pages are followed. Wikipedia Captchas appear on the account creation page. We note that textual reCaptchas, as of version 2.0, have been replaced with tasks such as image recognition \cite{sivakorn2016robot}, making them unlikely to encounter and collect.} collecting and hand-labeling test sets of 500 images each. We found that the trained Wikipedia and Facebook models achieving $>$~90\% recognition with synthetic data yielded practically zero breaking rates with real data.  We then tried using a model trained on one Captcha style to break another style and found that it nearly always failed as well.  We found that this was only partially caused by the non-overlapping latent variable domains (e.g., the distinct character ranges) for renderers of different styles.  For instance, one might expect the reCaptcha breaker to work on the visually similar Yahoo Captchas, but we found that this was not the case.

\begin{figure}
    \centering
    \def\arraystretch{2}
    \begin{tabular}{@{}p{0.3\columnwidth}p{0.325\columnwidth}p{0.3\columnwidth}@{}}
    \includegraphics[width=25mm]{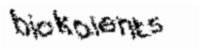} & \includegraphics[width=25mm]{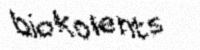} & \includegraphics[width=25mm]{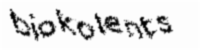}\\
    \includegraphics[width=25mm]{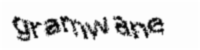} & \includegraphics[width=25mm]{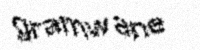} & \includegraphics[width=25mm]{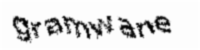}\\
    \includegraphics[width=25mm]{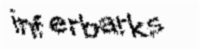} & \includegraphics[width=25mm]{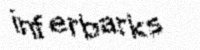} & \includegraphics[width=25mm]{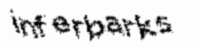}\\
    \includegraphics[width=25mm]{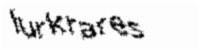} & \includegraphics[width=25mm]{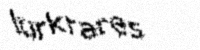} & \includegraphics[width=25mm]{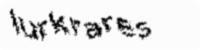}\\
    \includegraphics[width=25mm]{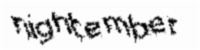} & \includegraphics[width=25mm]{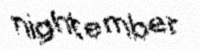} & \includegraphics[width=25mm]{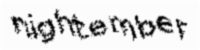}\\
    \includegraphics[width=25mm]{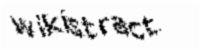} & \includegraphics[width=25mm]{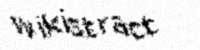} & \includegraphics[width=25mm]{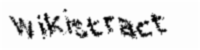}\\
    \end{tabular}

    \caption{Synthetic data from the Wikipedia generative model (left) are recognized correctly whereas even perceptually subtle changes such as adding per-pixel white noise with $\sigma = 5$ (middle) and $\epsilon_{kerning}$ modified by just one pixel (right) result in severely degraded recognition rates. The overall recognition rates for the test groups from which these samples are taken are 93.6\% (left), 24.0\% (middle) and 65.2\% (right). Note that the middle and right columns do get recognized correctly with the robust Wikipedia model.}
    \label{fig:brittleness}
\end{figure}

To investigate, we performed experiments where we constructed test Captchas that the trained networks cannot recognize despite being perceptually indistinguishable from Captchas from the original generative model.  We found that we could more-or-less arbitrarily degrade test performance by shifting the test data in either of two ways away from the original synthetic data (Figure~\ref{fig:brittleness}, left).  In the first (Figure~\ref{fig:brittleness}, middle), we corrupted the image by subtle additive noise which shifts each Captcha a small, imperceptible Euclidean distance from its original position.  This causes our Captcha breaking networks to exhibit the kind of brittleness well known to be a problem for deep neural network classifiers \cite{szegedy2013intriguing}.  In the second (Figure~\ref{fig:brittleness}, right), by changing the generative model of the test data relative to the training data, even in ways that are arguably below human ability to perceive, we were also able to cause test performance to degrade.  This is the kind of model misspecification that has been discussed in the Bayesian inference literature \cite{gelman2013philosophy}.
\begin{figure*}
    \centering
    \includegraphics[width=\textwidth,trim={2mm 0 2mm 0},clip]{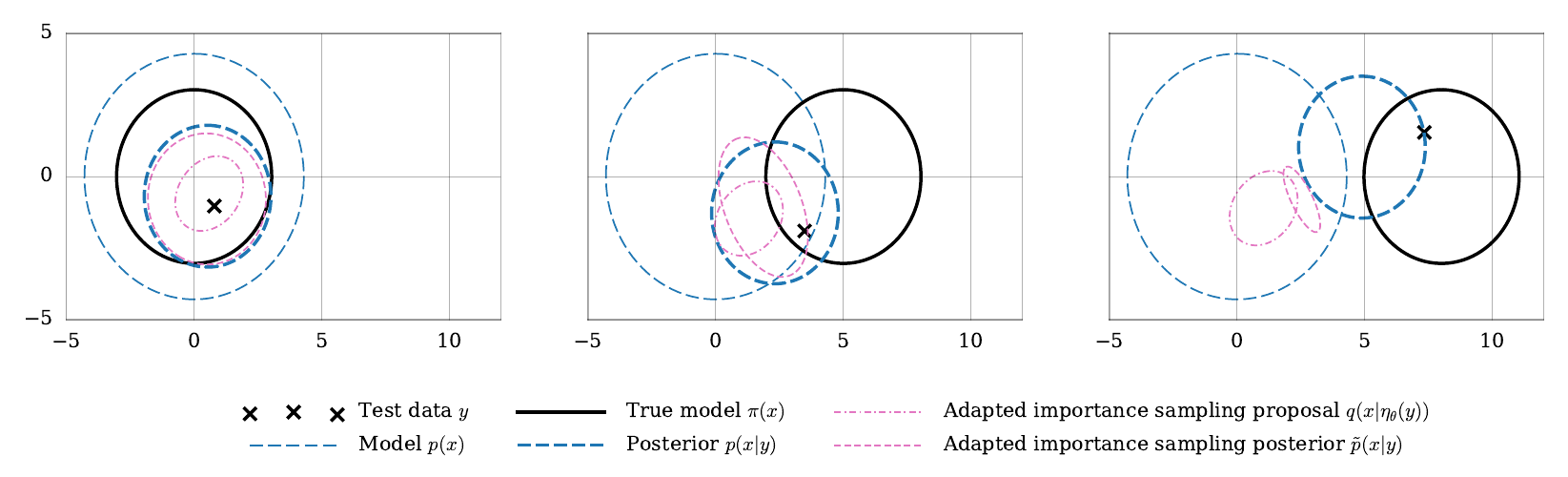}
    \caption{Illustration of model mismatch. Left: The model encompasses the true data distribution; Middle: the model partially matches the true data distribution; Right: the model is completely mismatched to the true data distribution.}
    \label{fig:connections/model-mismatch/gaussian}
\end{figure*}

Inspired by the success of \citet{jaderberg2014synthetic}, we attacked these problems by improving our synthetic training data generation.  In particular, we developed a substantially more flexible generative model using the elastic displacement fields introduced by \citet{simard2003}, effectively forcing the neural network to generalize over a greater variation than that exhibited by ground-truth labeled test data from the wild. These improved generative models have been observed to be robust to the subtle modifications that we report in Figure 2. The results we obtained are encouraging, achieving 81\% and 42\% recognition rates on real Wikipedia and Facebook Captchas respectively. In both cases our robust results, arrived at by improving the quality of the synthetic data generator, have performance comparable (in the case of Wikipedia, superior) to traditional vision pipelines, and are significantly higher than the 1\% recognition threshold suggested to deem a deployed Captcha system broken \cite{bursztein2011text}. 

\section{Discussion and connections to model-based Bayesian reasoning}
\label{sec:connections}

In order to explore some of the factors that cause the brittleness of the neural network performance that we have just reported, we draw a connection between Bayesian model mismatch and out-of-sample generalization failure of neural network and other regressors when tested on data that is different to that used for training.
%Following this, in this section, we also show that training neural networks in this way produces highly efficient proposal distributions for importance sampling.

As a prerequisite to this, we review importance sampling \citep{doucet2009tutorial}, the approximate probabilistic inference algorithm that most naturally 
corresponds to the kind of inference our trained neural networks allow us to do.
Given a joint distribution $p(x, y)$ and a user-specified proposal distribution $q(x \given y)$, importance sampling allows us to approximate the posterior distribution $p(x \given y)$ and expectations of arbitrary functions $f$ under it
\begingroup
\addtolength{\jot}{-2mm}
\begin{align}
    p(x \given y) &\approx \sum_{m = 1}^M W_m \delta(x - x^{(m)}) \label{eq:algorithm/is/posterior}\\
    \E_{p(x | y)}[f] &\approx \sum_{m = 1}^M W_m f(x^{(m)})\;. \label{eq:algorithm/is/expectation}
\end{align}
\endgroup
This is done by generating $M$ weighted samples $\{(w_m, x^{(m)})\}_{m = 1}^M$
\begingroup
\addtolength{\jot}{-2mm}
\begin{align}
    x^{(m)} &\sim q(x \given y) & m &= 1, \dotsc, M \\
    w_m &= p(x^{(m)}, y) / q(x^{(m)} \given y) & m &= 1, \dotsc, M \label{eq:weights}\\
    W_m &= w_m / \sum_j w_j & m &= 1, \dotsc, M\;.
\end{align}
\endgroup
Note that importance sampling is generally inefficient unless the proposal distribution is well-matched to the target distribution in the sense that it ``overlaps'' the target, and is extremely efficient if it matches exactly.

\subsection{Bayesian model misspecification}
\label{sec:connections/model-mismatch}

We illustrate the effects of mismatch between synthetic and real data in terms of Bayesian model misspecification using a simpler experiment (Figure~\ref{fig:connections/model-mismatch/gaussian}), highlighting conceptually what we believe to be happening.
Let $\pi(x, y)$ be the true data generating distribution and $p(x, y)$ a model, where
\begingroup
\addtolength{\jot}{-2mm}
\begin{align}
  \pi(x) &= \Gauss(x \given \mu_{\pi}, \Sigma_{\pi}) \\
  \pi(y \given x) &= \Gauss(y \given x, \Sigma) \\
  p(x) &= \Gauss(x \given \mu_p, \Sigma_p) \\
  p(y \given x) &= \Gauss(y \given x, \Sigma)\;.
\end{align}
\endgroup
We will use the mismatch between the distributions $\pi(x)$ and $p(x)$ as an illustrative proxy to the mismatch of the joint distributions $\pi(x, y)$ and $p(x, y)$.

The marginal $p(x)$ of the model distribution $p(x, y)$ is shown in Figure~\ref{fig:connections/model-mismatch/gaussian} as a thin blue dashed ellipse which covers 99\% of its probability mass.
We draw a data point $y$ from this model by first drawing $x$ from $p(x)$ and then drawing $y$ from $\pi(y \given x)$ where $\Sigma_p = 2I$, $\mu_p = [0, 0]^\mathsf{T}$ and $\Sigma = I$.

The marginal $\pi(x)$ of the true data generating distribution $\pi(x, y)$ is shown in Figure~\ref{fig:connections/model-mismatch/gaussian} as a black solid ellipse.
A typical data point $y$ is drawn by first drawing $x$ from $\pi(x)$ and then drawing $y$ from $\pi(y \given x)$ where $\Sigma_\pi = I$ and $\mu_{\pi}$ is $[0, 0]^\mathsf{T}, [5, 0]^\mathsf{T}$ and $[8, 0]^\mathsf{T}$ from left to right.

% We define $p(x, y) = p(y \given x) p(x)$ in terms of the prior $p(x)$ and the likelihood $p(y \given x)$:
% \begin{align}
%     x &\sim \Gauss(\mu_0, \Sigma_0) \\
%     y | x &\sim \Gauss(x, \Sigma) 
% \end{align}
% where the prior mean $\mu_0 = [0, 0]^T$, prior covariance $\Sigma_0 = 2I$, and likelihood covariance $\Sigma = I$.
% The prior $p(x)$ in the generative model is shown in Figure~\ref{fig:connections/model-mismatch/gaussian} as a dark blue ellipse.

Such a model has a posterior
\begingroup
\addtolength{\jot}{-1mm}
\begin{align}
  p(x \given y) &= \Gauss(x \given \mu_{\text{post}}, \Sigma_{\text{post}}) \\
  \Sigma_{\text{post}} &= (\Sigma_p^{-1} + \Sigma^{-1})^{-1} \\
  \mu_{\text{post}} &= \Sigma_{\text{post}}(\Sigma_p^{-1} \mu_p + \Sigma^{-1} y)\;,
\end{align}
\endgroup
which is shown in Figure~\ref{fig:connections/model-mismatch/gaussian} as a thick blue dashed ellipse.

% Such a model has posterior $p(x | y) = \Gauss(\mu_1, \Sigma_1)$ with $\Sigma_1 = (\Sigma_0^{-1} + \Sigma^{-1})^{-1}$ and $\mu_1 = \Sigma_1(\Sigma_0^{-1} \mu_0 + \Sigma^{-1} y)$.
% The posterior is shown in Figure~\ref{fig:connections/model-mismatch/gaussian} as a dark green ellipse.

Using a procedure similar to the one described in Section~\ref{sec:captcha-breaking}, we generate training data $\{(x^{(n)}, y^{(n)})\}$ from the model $p(x, y)$ and use it to train a neural network mapping from $y$ to importance sampling proposal parameters $(\mu_q, \Sigma_q) := \eta_{\theta}(y)$.
The resulting proposals generated from such a proposal distribution $q(x \given \eta_{\theta}(y)) := \Gauss(x \given \mu_q, \Sigma_q)$ are shown in Figure~\ref{fig:connections/model-mismatch/gaussian} as magenta dash-dotted ellipses.  Remember that $\mu_q$ and $\Sigma_q$ are functions of $y$ computed by the trained neural network regressor.

If we then draw $M = 1000$ samples from this proposal distribution by repeatedly running the trained neural network forward and weight the resulting samples according to the importance sampling scheme in the beginning of Section~\ref{sec:connections}, we arrive at approximations to the model-based posterior mean and covariance:
\begingroup
\addtolength{\jot}{-1mm}
\begin{align}
  \tilde \mu_M &\approx \E_{p(x \given y)}[x] \\
  \tilde \Sigma_M &\approx \E_{p(x \given y)}\left[\left(x -  \E_{p(x \given y)}[x]\right)\left(x -  \E_{p(x \given y)}[x]\right)^\mathsf{T}\right].
\end{align}
\endgroup
The distribution $\tilde p(x \given y) := \Gauss(x \given \tilde \mu_M, \tilde \Sigma_M)$ is shown in Figure~\ref{fig:connections/model-mismatch/gaussian} as a magenta dashed ellipse.

Now consider the three scenarios in Figure~\ref{fig:connections/model-mismatch/gaussian}, in which the difference between the true data generating distribution, illustrated by its marginal $\pi(x)$, and the model $p(x)$ is progressively increased from left to right.
As the true data generating distribution $\pi(x)$ moves further away from our model $p(x)$ we see that we get, for a fixed computational budget of $M = 1000$ samples, progressively worse estimates $\tilde p(x \given y)$ of $p(x \given y)$  (Figure~\ref{fig:connections/model-mismatch/gaussian}, middle and right).
What is happening here is that the neural network, at training time, learns to invert the model $p(x,y)$ from samples drawn from it.
In Figure~\ref{fig:connections/model-mismatch/gaussian} (left), when the model overlaps the true data generative process, the neural network sees examples of $x$ and $y$ pairs that are representative of the true data generating mechanism and then, given sufficient capacity in terms of neural architecture and training time (remembering that we have access in this setting to infinite training data), can almost certainly learn a mapping that solves the task of predicting $x$ given $y$.
If the model is slightly misspecified then the number of training examples in the domain of the true model might be small and as such we might not expect good generalization performance.
When there is high model misspecification (Figure~\ref{fig:connections/model-mismatch/gaussian}, right) the neural network will simply never see training examples that look like the true data, and, as such, will produce mostly spurious regression results leading to unhelpful proposal distributions.

This experiment graphically illustrates the kinds of problems that can arise from model misspecification. What it indicates is that if we are going to use synthetic data to train a neural network regressor we should ensure that our synthetic data generator is ideally as close as possible to the true data generation process and that mismatch from the true data in terms of broadness (e.g., the Gaussian example in Figure~\ref{fig:connections/model-mismatch/gaussian} (left), in which $\mu_{\pi}$ and $\mu_p$ match but $\Sigma_{\pi}$ and $\Sigma_p$ do not) is more tolerable and in fact preferable to a  perceptually indistinguishably miscalibrated model (e.g. the phenomenon illustrated in Figure~\ref{fig:brittleness} and described in Section~\ref{sec:experiments/robustness}).
% We contend that this type of model mismatch is what caused the brittleness we discovered in our trained neural networks and illustrate in Figure~\ref{fig:brittleness}.
We conjecture that the latter is what caused the brittleness we discovered in our trained neural networks and illustrate in Figure~\ref{fig:brittleness}.

This intuition guided our decision to broaden our synthetic data generator by adding the displacement fields of \citet{simard2003} in Section~\ref{sec:experiments/robustness}, leading to significant improvements to robustness evidenced by the improved real-data results we obtained. This, we believe, accounts for the fact that our Captcha generator is not likely to capture all details of the true generative model such as subtle font differences.

\subsection{Inference}
\label{sec:connections/inference}

A corollary to the Bayesian inference interpretation of training a neural network on synthetic data is that the resulting neural network can be used for approximate inference in the probabilistic model $p(x,y)$ corresponding to the synthetic training data generator.  

Let the importance sampling proposal distribution be factorized as $q(x \given y) = \prod_{t = 1}^T q_t(x_t \given x_{1:t - 1}, y)$.
If we consider the individual time-dependent softmax layers of the Captcha-solving neural network to be probabilities of a  proposal distribution $q_t(x_t \given x_{1:t - 1}, y)$, we can adopt an alternative way of writing our loss in \eqref{eq:nn/loss} as
\begingroup
\addtolength{\jot}{-1mm}
\begin{align}
    \mathcal L(\theta) &= \frac{1}{N}\sum_{n = 1}^N \left[-\sum_{t = 1}^T \log\left( [\eta_{\theta, t}(y^{(n)})]_{x_t^{(n)}} \right)\right] \nonumber\\
    &= \frac{1}{N}\sum_{n = 1}^N \left[ -\sum_{t = 1}^T \log q_t( x_t^{(n)} \given x_{1:t - 1}^{(n)}, y ) \right] \nonumber\\
    &= \frac{1}{N}\sum_{n = 1}^N \left[ - \log \left( \prod_{t = 1}^T q_t( x_t^{(n)} \given x_{1:t - 1}^{(n)}, y ) \right)\right] \nonumber\\
    &= \frac{1}{N}\sum_{n = 1}^N \left[ -\log q(x^{(n)} \given \eta_{\theta}(y^{(n)})) \right]\;. \label{eq:connections/loss}
\end{align}
\endgroup
The loss in \eqref{eq:connections/loss} can be viewed as a Monte Carlo approximation of an expectation over a function under the joint distribution $p(x, y)$ of the synthetic data, which, following \citet{paige2016inference}, can be shown to be the Kullback-Leibler divergence between the proposal and the posterior averaged over all possible datasets
\begingroup
\addtolength{\jot}{-1mm}
\begin{align}
    &\E_{p(x, y)}[-\log q(x \given \eta_{\theta}(y))] \nonumber \\
    &= \int_{\mathcal Y} \int_{\mathcal X} p(x, y) (-\log q(x \given \eta_{\theta}(y))) \,\mathrm dx \mathrm dy \nonumber \\
    &= \int_{\mathcal Y} p(y) \int_{\mathcal X} p(x \given y) \log \frac{p(x \given y)}{q(x \given \eta_{\theta}(y)))} \,\mathrm dx \mathrm dy + \text{const.} \nonumber\\
    &= \E_{p(y)}[\KL{p(x \given y)}{q(x \given \eta_{\theta}(y))}] + \text{const.}
\end{align}
\endgroup
Hence, minimizing \eqref{eq:connections/loss} is also known as importance sampling proposal adaptation.

%Viewed this way, we can not only use our neural network's output as proposal distributions for an importance sampling algorithm, we have also demonstrated that doing so generates extremely efficient proposal distributions over many new data points, in this case, Captchas.

\begin{figure*}
  \centering
  \caption{Posteriors of real Facebook and Wikipedia Captchas. Conditioning on each Captcha, we show an approximate posterior produced by a set of weighted importance sampling particles $\{(w_m, x^{(m)})\}_{m = 1}^{M = 100}$.}
  \label{fig:results/posterior-histograms}
  \vspace{4mm}
  \begin{tabularx}{\textwidth}{@{}XXXXX@{}}
  \includegraphics[width=0.19\textwidth]{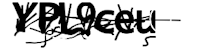} & 
  \includegraphics[width=0.19\textwidth]{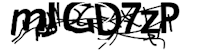} &
  \includegraphics[width=0.19\textwidth]{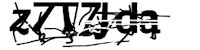} &
  \includegraphics[width=0.19\textwidth]{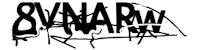} &
  \includegraphics[width=0.19\textwidth]{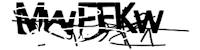} \\\vspace{-6mm}
  \includegraphics[width=\textwidth,trim={4mm 0 0 0},clip]{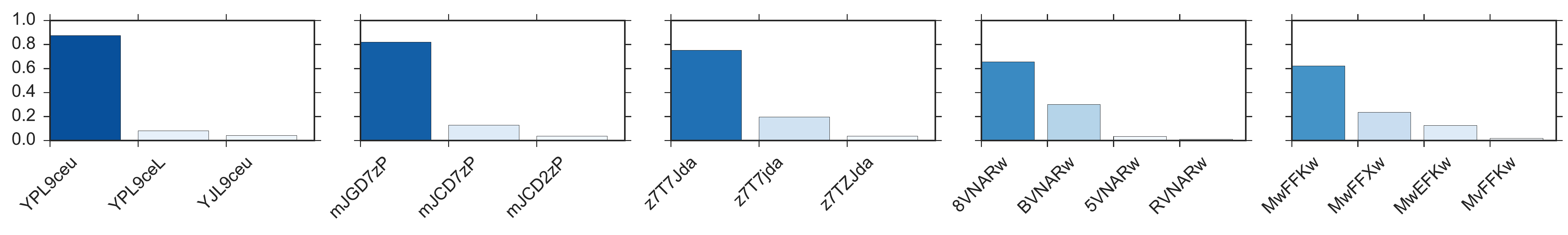}\\

  \includegraphics[width=0.19\textwidth]{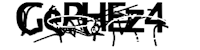} & 
  \includegraphics[width=0.19\textwidth]{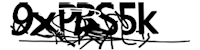} &
  \includegraphics[width=0.19\textwidth]{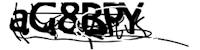} &
  \includegraphics[width=0.19\textwidth]{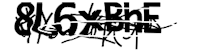} &
  \includegraphics[width=0.19\textwidth]{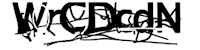} \\\vspace{-6mm}
  \includegraphics[width=\textwidth,trim={4mm 0 0 0},clip]{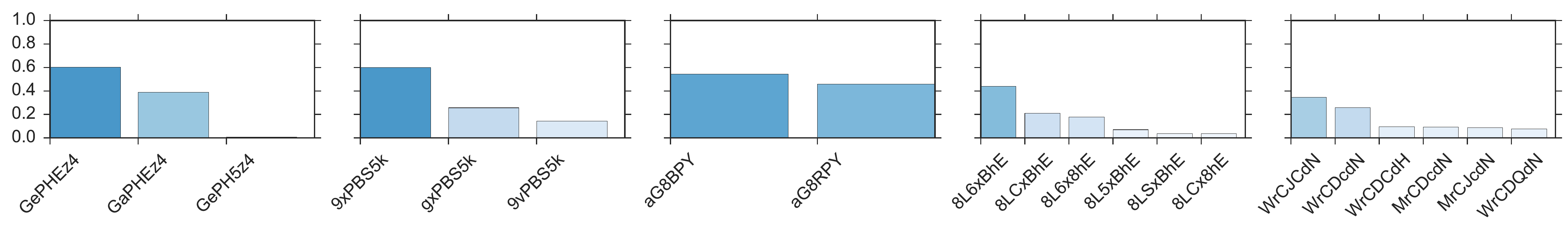}\\

  \includegraphics[width=0.19\textwidth]{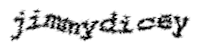} & 
  \includegraphics[width=0.19\textwidth]{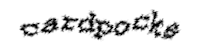} &
  \includegraphics[width=0.19\textwidth]{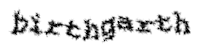} &
  \includegraphics[width=0.19\textwidth]{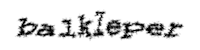} &
  \includegraphics[width=0.19\textwidth]{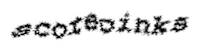} \\\vspace{-6mm}
  \includegraphics[width=\textwidth,trim={4mm 0 0 0},clip]{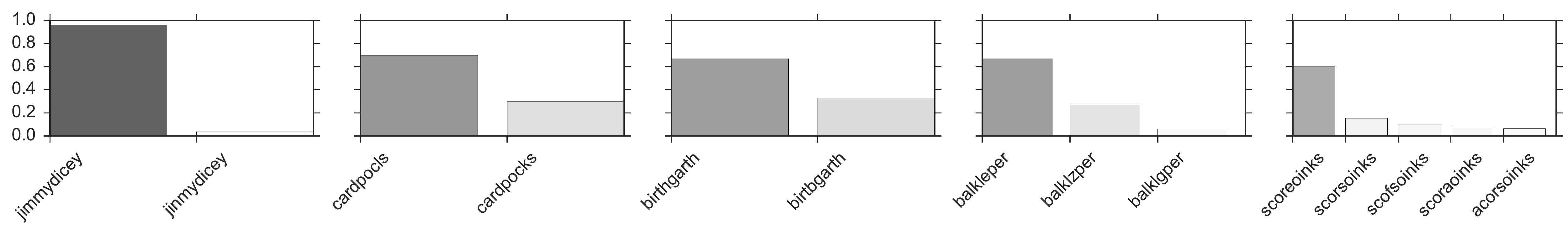}\\

  \includegraphics[width=0.19\textwidth]{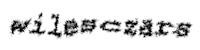} & 
  \includegraphics[width=0.19\textwidth]{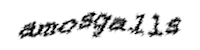} &
  \includegraphics[width=0.19\textwidth]{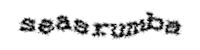} &
  \includegraphics[width=0.19\textwidth]{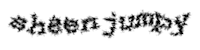} &
  \includegraphics[width=0.19\textwidth]{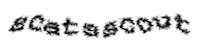} \\\vspace{-6mm}
  \includegraphics[width=\textwidth,trim={4mm 0 0 0},clip]{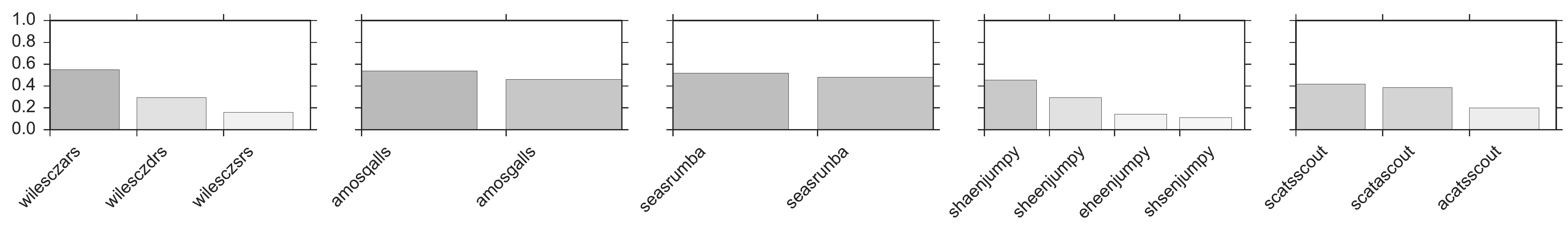}\\
  \end{tabularx}
  \vspace{-3mm}
\end{figure*}

Running a neural network trained using synthetic data and this common loss on an input $y$ actually produces efficient proposal distribution parameters $\eta_{\theta}(y)$. By running the neural network $M$ times given the same input and subsequently weighting the sampled $x$ values according to \eqref{eq:weights}, we obtain an approximate posterior distribution (Figure~\ref{fig:results/posterior-histograms}).
We note that, in the case of Captchas, we must use a likelihood based on approximate Bayesian computation (ABC) \cite{wilkinson2013approximate} instead of the intractable $p(y \given x)$ in order to calculate the weight in \eqref{eq:weights}.

Accounting for uncertainty is a principal benefit of model-based inference and is particularly useful when there is actual ambiguity in $y$ as in Figure~\ref{fig:results/posterior-histograms}.

%These posterior histograms allow interpretable Captcha solving with human-like uncertainties.
%This would not have been possible with importance sampling with proposals from $p(x)$ since the search space is too large. This is in contrast to related work in probabilistic programming based Captcha solving work by \citet{mansinghka2013approximate}. That work obtains only point estimates whereas our approach provides posterior estimates.

\section{Conclusion}
\label{sec:conclusion}

What is remarkable about the natural scene text recognition results of \citet{jaderberg2014synthetic,jaderberg2016reading} is that they show generalization from synthetic data, to the degree that one could argue that their result
is actually a generative modeling triumph.  Our results showing improved robustness of Wikipedia- and Facebook-style Captcha-breaking stem likewise from focusing on the synthetic data generative model.
%In a more general note, we stress that these results apply beyond the domain of Captchas.
In addition to being usefully prescriptive, our point that training neural networks using synthetic data is equivalent to performing proposal adaptation for importance sampling inference in the synthetic data generative model sets an empirical cornerstone for future theory that quantifies and bounds the impact of model mismatch on neural network and approximate inference performance.

\section*{Acknowledgments}

Tuan Anh Le is supported by EPSRC DTA and Google (project code DF6700) studentships. Atılım Güneş Baydin and Frank Wood are supported under DARPA PPAML through the U.S. AFRL under Cooperative Agreement FA8750-14-2-0006, Sub Award number 61160290-111668. Robert Zinkov is supported under DARPA grant FA8750-14-2-0007.



%\bibliographystyle{IEEEtran}


% \end{thebibliography}

\setlength{\bibsep}{1pt}
\bibliographystyle{IEEEtranN}
\bibliography{captcha}

\end{document}